\documentclass[letterpaper, 10 pt, conference]{ieeeconf}

\IEEEoverridecommandlockouts
\usepackage{cite}
\usepackage{amsmath,amssymb,amsfonts}
\usepackage{algorithmic}
\usepackage{graphicx}
\usepackage{textcomp}
\usepackage{xcolor}
\usepackage{indentfirst}
\usepackage{booktabs}
\usepackage{soul}
\def\BibTeX{{\rm B\kern-.05em{\sc i\kern-.025em b}\kern-.08em
    T\kern-.1667em\lower.7ex\hbox{E}\kern-.125emX}}
\usepackage{mathtools}

\usepackage{csquotes}

\makeatletter
\let\NAT@parse\undefined
\makeatother
\usepackage{hyperref}

\title{\bf \LARGE Wearable Roller Rings to Augment In-Hand Manipulation through Active Surfaces\\
}
\author{Hayden M. Webb$^1$, Podshara Chanrungmaneekul$^1$, Shenli Yuan, and Kaiyu Hang$^1$
    \thanks{$^1$Department of Computer Science, Rice University, Houston, TX 77005, USA. This work was supported by US National Science Foundation award FRR-2240040.}%
}

\begin{document}

\maketitle

\begin{abstract}
In-hand manipulation is a crucial ability for reorienting and repositioning objects within grasps. The main challenges in this are not only the complexity of the computational models, but also the risks of grasp instability caused by active finger motions, such as rolling, sliding, breaking, and remaking contacts. 
This paper presents the development of the Roller Ring (RR), a modular robotic attachment with active surfaces that is wearable by both robot and human hands to manipulate without \textit{lifting a finger}.
By installing the angled RRs on hands, such that their spatial motions are not colinear, we derive a general differential motion model for manipulating objects. Our motion model shows that complete in-hand manipulation skill sets can be provided by as few as only $2$ RRs through non-holonomic object motions, while more RRs can enable enhanced manipulation dexterity with fewer motion constraints. Through extensive experiments, we test the RRs on both a robot hand and a human hand to evaluate their manipulation capabilities. We show that the RRs can be employed to manipulate arbitrary object shapes to provide dexterous in-hand manipulation.


\end{abstract}
\section{Introduction}
To physically engage robots with our daily tasks, evolving new modalities of in-hand manipulation skills is crucial for addressing problems associated with the traditional manipulation paradigms. In the past decades, extensive studies have worked towards improving in-hand manipulation by developing new hand designs, computational models, and more integrated systems. These not only mechanically provide better options, but also computationally and perceptually enhance their manipulation capabilities \cite{calandra2018more, bullock2011classifying, piazza2019century}. 

Typically, in-hand manipulation is performed using precision grasps due to its robust dexterity. However, such hand-object configurations come at the cost of stability and strength due to the small fingertip contacts. Moreover, due to the high degrees of freedom, precision grasp manipulation often relies on computationally complex, small-scale, and vulnerable non-holonomic movements. In contrast, using power grasps is more stable, but typically lacks the capability to perform relative motions between the object and hand. While in-hand manipulation, in essence, is about stably maintaining and moving contacts with the manipulated object, most traditional approaches fail to encompass these values.

\begin{figure}[t]
    \centering
    \includegraphics[width=1.0\linewidth]{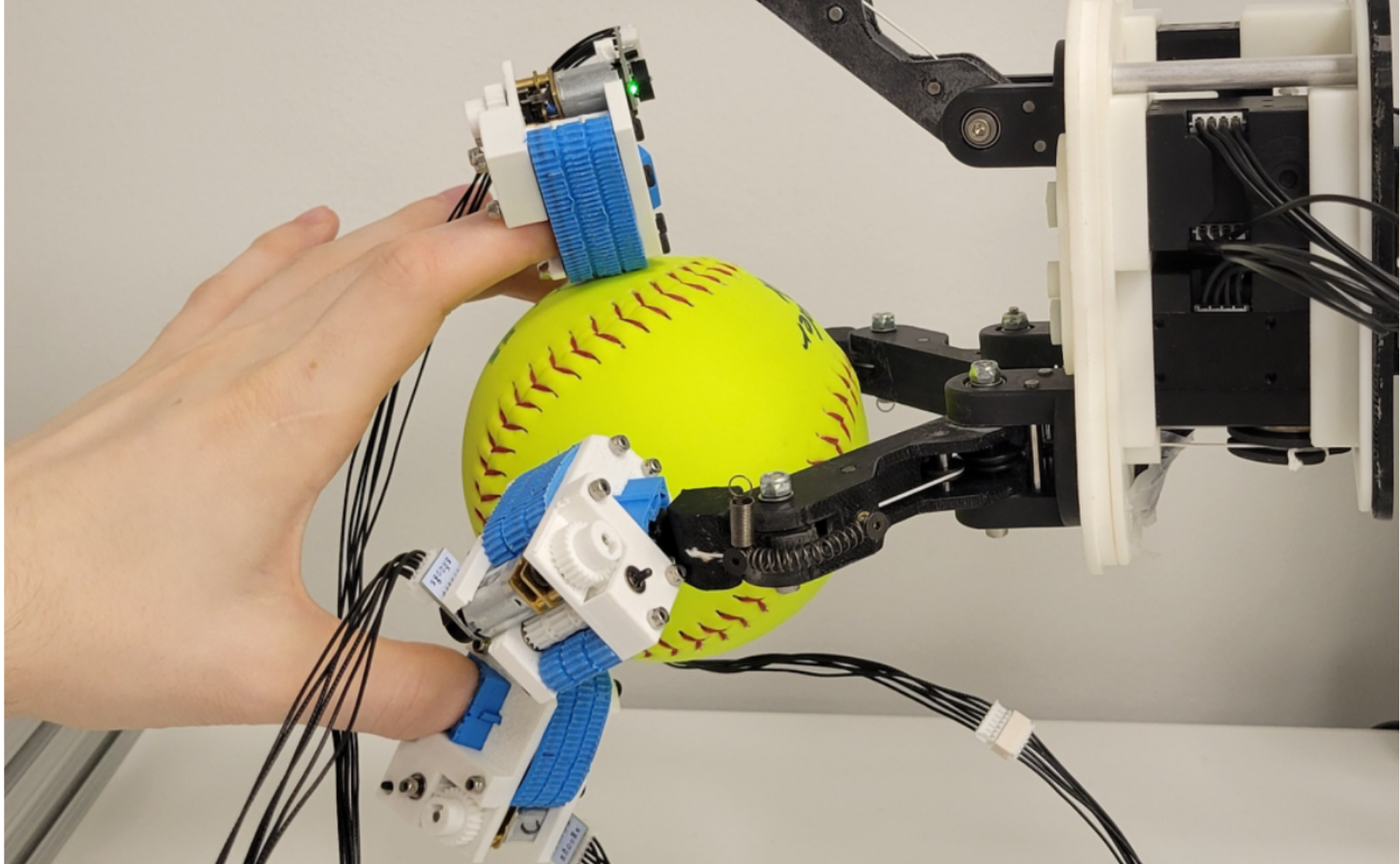}
    \caption{Four Roller Rings, affixed on both a robot hand and a human hand, that are co-grasping a softball. Configurations like these can enable in-hand human-robot co-manipulation for grasped objects.}
    \label{fig:reference}
\end{figure}

To this end, a number of novel hands have been developed to fill in the aforementioned gaps. Underactuated hands have been designed to enable passive stability maintenance, while requiring significantly fewer degrees of control to be regulated for in-hand manipulation \cite{hang2021manipulation}. Yet, such designs are limited to very small-scaled manipulation when utilizing manipulation techniques, such as finger-gaiting \cite{morgan2022complex}, due to their high complexity and lack of stability. Alternatively, compliant palm support-based manipulation has been explored to greatly expand manipulation ranges \cite{bhatt2022surprisingly, andrychowicz2020learning, pagoli2021palm}, however, without much consideration for their manipulation precision. Moreover, techniques and hands such as these require non-holonomic movements in their planning. This is due to the potential for unstable grasps or collisions along the direct manipulation path between arbitrary positions, or the lack of guarantees for orthogonal axes of rotations (Further discussion of this can be found in Sec.~\ref{sec:motionModel}). This necessitates detours to ensure stability and movement while guaranteeing a full manipulation solution. Despite this, these hands and techniques do guarantee full movement solutions for almost all in-hand manipulation.

Alternatively, the concept of active surfaces \cite{datseris1985principles}, which actively moves a grasp contact by directly translating the contact on the hand, has created a new modality for in-hand manipulation. Hands developed with active surfaces are capable of performing dexterous manipulation \emph{without lifting a finger} and, in theory, can translate the contacts without limitations. However, this has not proven to be the case for previous designs utilizing active surfaces. For rigid hands, motions actuated by active surfaces need to be precisely controlled to maintain the grasp stability \cite{yuan2020design,yuan2020grasper,rgv3}. The BACH hand further combined mechanical compliance and active surfaces into a hand design \cite{cai2023power}, enabling unprecedented in-hand manipulation without requiring complex computational models or control schemes while still garnering precision grasp robustness and power grasp stability. However, a major limitation of the BACH and other similar systems is that the benefits from active surfaces cannot be directly transferred to other hands. Likewise, algorithms for these designs are integrated into the system's design directly rather than being generally integratable across designs. Thus, these designs and their solutions only exist in an \textit{ad hoc} manner and cannot be transferred to other existing hands.

To this end, we, in this work, propose the design of the Roller Rings (RR) (exemplified in Fig.~\ref{fig:reference}) -- a new form of wearable devices that can enable and augment active surface-based in-hand manipulation for \emph{any hand} due to its modularity. In brief, this work provides:
\begin{enumerate}
    \item The world's first wearable device for manipulation augmentation that is low-cost and easy to fabricate;
    \item A complete manipulation solution through differential manipulation motions (Sec.~\ref{sec:motionModel});
    \item A manipulation device attachable to any robot and human hand while not changing the existing capability of the original hands.
\end{enumerate}

We discuss related research in Sec.~\ref{sec:relatedWork}, then describe the design principles for the Roller Rings in Sec.~\ref{sec:design}. In Sec.~\ref{sec:motionModel}, we derive the differential motion model for the RR-based in-hand manipulation. Real-world experiments with both robot and human hands are discussed in Sec.~\ref{sec:experiments}. Finally, Sec.~\ref{sec:conclusion} concludes and discusses the future work.

\section{Related Work}
\label{sec:relatedWork}

In the past decades, a myriad of robot hands have been developed to carry out dexterous manipulation~\cite{piazza2019century}. The Stanford/JPL hand~\cite{salisbury1983kinematic} and the Utah/MIT hand~\cite{jacobsen1986design}, for example, are among the earliest hands with in-hand manipulation capabilities. Because a number of robot hands are inspired by human hands, common biomimetic manipulation techniques such as finger-gaiting became a natural choice for robot manipulation~\cite{leveroni1996reorienting, han1998dextrous, morgan2022complex, ma2014underactuated}. During finger-gaiting, robot fingertips will sequentially break and remake contacts, allowing fingers to ''walk" along the grasped object to perform manipulation. However, fingertip movements are limited by grasp stability and small motion scales. Differing from finger-gaiting approach, rolling manipulation~\cite{montana1988kinematics,bicchi1995dexterous,choudhury2002rolling} allows the grasped object to be moved in-hand without breaking contacts, enabling simpler controls and more stable manipulation. However, this approach is still challenged by the small scale of in-hand object motion ranges. Moreso, these approaches are similar in that their respective movement schemes are nominally non-holonomic.

\subsection{Rolling Manipulation and Active Surfaces}
 Robot grippers with active surfaces~\cite{tincani2012velvet,ma2016primitives,gomez2021adaptive,yuan2020design,yuan2020grasper,rgv3,cai2023power} are specifically designed to perform rolling manipulation. These grippers have shown their unique abilities in various in-hand manipulation tasks, as they allow the grasped object to be manipulated without the need to lift fingers during manipulation. The velvet finger~\cite{tincani2012velvet}, for example, utilizes a set of actuated fingers inlaid with active surfaces to perform both rotational and translational manipulation.  Additionally, the Roller Graspers~\cite{yuan2020design,yuan2020grasper,rgv3} have active surfaces designed into their fingertips to dexterously grasp and manipulate objects with the combination of roll and pitch movements. A more recent work~\cite{cai2023power} demonstrated that active surfaces and mechanically compliant grasps can be used to dexterously manipulate objects without any complex computational models or control schemes non-holonomically. However, these designs do not generalize to, or enhance, the capabilities of any existing hand designs. Thus, existing hands cannot benefit manipulation-wise without modifying their existing capabilities or completely redesigning the system. In this work, we address these issues by creating an active surface-based manipulator that can be easily affixed to existing systems to provide these capabilities (See Sec.\ref{sec:design}) without compromising the existing system.

\subsection{Algorithmic Approaches to Active Surfaces}
Computational models have been derived to search for the optimal combinations of active surface motions to enable in-hand manipulation. For example, \cite{xie2023parallel} samples certain grasps using object mesh models to generate object rotations. Using this approach, the sum of combinations of angular velocities is standardized to actuate the in-hand motions of the grasped object. \cite{isobe2023vision,isobe2023contact} used a vision-based control system to determine the geometric information of an object in the next frame and calculated the necessary control commands of the active surface motions to achieve it. Additionally, employing control over both the belts and fingers greatly increases the flexibility of the active surface-based manipulation, while at the cost of more complex hand-object models \cite{yuan2020grasper}. However, these existing algorithms are all designed in an \emph{ad hoc} manner to only work for their respective designs. Therefore, existing hands are not able to directly benefit from these approaches. To this end, this work presents a generalized Motion Model (See Sec. \ref{sec:motionModel}) to remediate this discrepancy.

\subsection{Human Augmentation}
Numerous augmentation devices have been designed utilizing the same concepts as robotic hands to mitigate the gap in in-hand manipulation for users~\cite{massa2002design, zhao2006prosthetic, cheng2023development}. Thus, these approaches are nominally focused on recreating power and precision grasps in their designs. The Yale Multigrasp Hand \cite{2016BelterMultigrasp}, for example, uses a series of body-powered cables to recreate different forms of grasps to facilitate manipulation. Likewise, myoelectric augmenters, like in \cite{leddy2018Prosthetic}, utilize a set of actuated cables that drive underactuated fingers through signals sent by the body. Both devices' fingers can be used in conjunction to manipulate a grasped object. However, because these designs utilize biomimetic power and precision grasps, augmentations like these are naturally lackluster at performing complex in-hand manipulation tasks. Moreso, as often seen in nature, these manipulation paths are non-holonomic and often require detours during manipulation due to the lack of stable direct control in this grasp formation. Thus, this work provides the framework for a manipulation device that can be affixed to any biomimetic hand to augment their in-hand manipulation capabilities (See Sec.~\ref{sec:experiments}).
\section{Design}
\label{sec:design}

\begin{figure*}[t]
    \centering
    \includegraphics[width=1.0\linewidth]{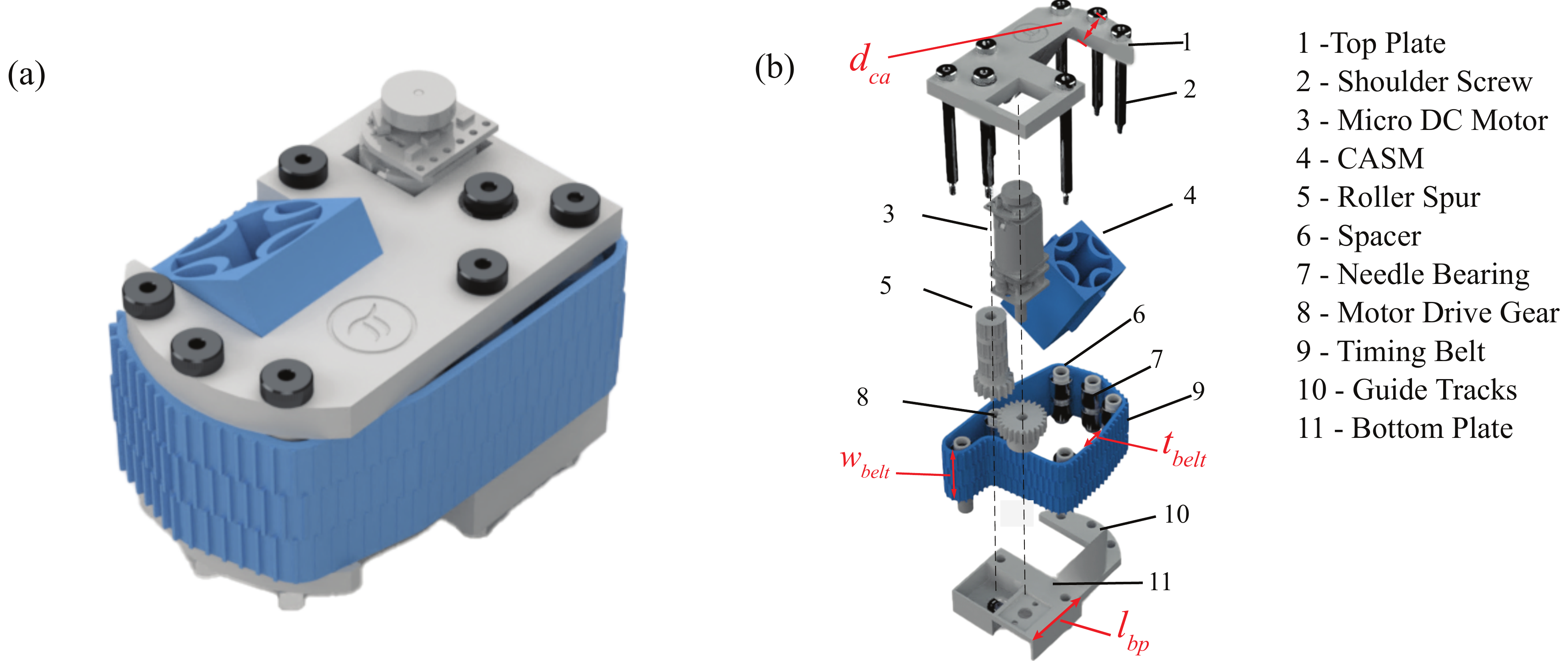}
    \caption{CAD Model of the Roller Ring: Exploded view of the device; Dimensions of each specified variable can be found in Table~\ref{tab:model specifications}}
    \label{fig:design parts} 
\end{figure*}

While active surfaces have shown promising capabilities for in-hand manipulation tasks, previous active surface manipulators are limited to their underlying hand designs. In this work, we aim to improve the manipulation capabilities of existing hands through wearable and modular RRs. With the integration of the RRs, hands that are designed for various grasping and manipulation purposes with different form factors will have the ability to manipulate grasped objects without breaking contacts (i.e. lifting their fingers). To achieve this goal, we propose the following design principles:

\begin{enumerate}
    \item The design should provide a full spatial manipulation capability, i.e., moving a grasped object from an arbitrary initial pose to an arbitrary target pose;
    \item While compactness is not a main design objective, the proof-of-concept should still be sufficiently compact to exemplify the RRs core capabilities
    \item The RRs should be customizable in order for them to adapt to a wide range of hands and fingers.
\end{enumerate}

\subsection{Components}

A CAD model of the RR is presented in Fig.~\ref{fig:design parts} and its dimensions can be found in Table~\ref{tab:model specifications}. Each RR variant has a one degree of freedom (DoF) active surface that is achieved using a timing belt routed through a series of needle bearings. The single DoF design is used as opposed to the 2-DoF (pivot and roll) design in~\cite{yuan2020design,yuan2020grasper} to minimize the form factor of the RR according to the guiding principles. Without the additional pivot joint, multiple RRs must be used simultaneously to generate a net differential motion (see Sec.~\ref{sec:motionModel}). The plates and gears of the RR were 3D-printed using Polylactic Acid (PLA, Bambu PLA Basic) at 30\% infill and the actuated timing belt and CASM were 3D-printed using Thermoplastic Polyurethane (TPU, NinjaTek Chinchilla 75A).

The belt is driven by a 12V N20 Micro DC Motor along a roller spur. It was routed, as seen in Fig.~\ref{fig:design parts}, to minimize the footprint of the whole RR model while satisfying the following constraints: \textbf{(1)} The timing belt forms a convex active surface where the object is contacted, and \textbf{(2)} The belt has enough contact area with the roller spur to avoid stalling. The surface pattern of the outer side of the timing belt is designed to mate with the roller spur's teeth, which also doubles as a high-friction surface for grasping and manipulation. Concurrently, it has grooves on the inner side to fit onto the needle bearings and constrains the lateral motion, similarly to~\cite{cai2023power}.

\begin{figure}
    \centering
    \includegraphics[width=1.0\columnwidth]{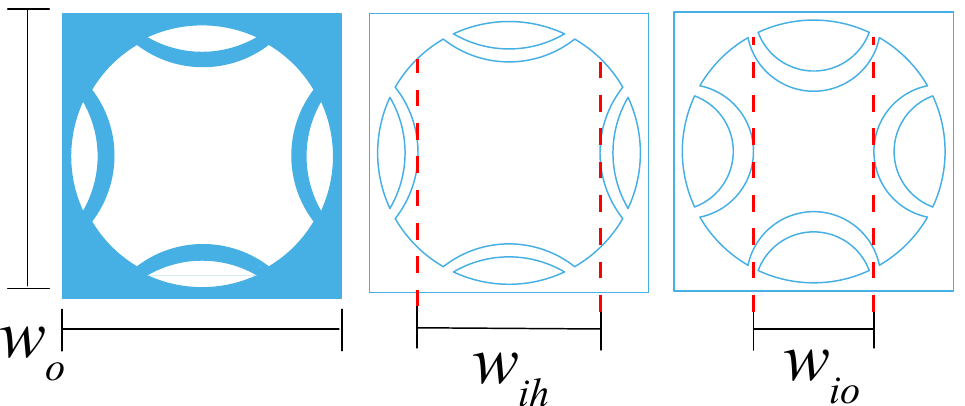}
    \caption{CASM Variants for Roller Ring (Left to Right: Base Model, Human, Yale Model O) Variable dimensions for variants can be found in Table~\ref{tab:model specifications}.}
    \label{fig:CASM variants}
\end{figure}

The RR can be mounted on various hands through a customizable Conformable Affixing Sleeve Module (CASM) mounted on the guide track of the RR (Fig.~\ref{fig:design parts}(10)). To ensure that the device was generally wearable by any hand, we employed the use of an inverted quatrefoil as seen in Fig.~\ref{fig:design parts}(4). This was chosen due to its ability to correctly deform and resist plastic deformation under compression due to the filament and geometry's innate compliance~\cite{azmi2020quatrefoil}. This high level of compliance allowed for the CASM to be utilized in both the robot and human tests (Fig.~\ref{fig:CASM variants}). The CASM variants were designed following these constraints to ensure correct affixment: \textbf{(1)} The outer width of the CASM ($w_o$) should be at least $2$ mm greater than the greatest width of the attachment point for the set of utilized systems to ensure space for fin deformation, \textbf{(2)} the quatrefoil's fins has to be at least $1$ mm thick to prevent plastic deformation, and \textbf{(3)} the inner width ($w_i$) has to be at least $2$ mm less than the attachment point to allow for quatrefoil deformation around the object. We chose a $w_o$ of $17$ mm, due to the human finger having the largest width at $15$ mm of the set that the CASM would be affixed on in our experiments, and thus being the maximum diameter the CASM can cover. Further demonstration of the CASM can be found in Sec.~\ref{sec:experiments}.

\subsection {Angled Active Surface}

The design principle (1) specifies that the RR be capable of performing full spatial manipulation. Considering parallel finger configuration is very common in multi-fingered grasping and manipulation, we designed the active surface to be angled from the axis of the finger the RR is mounted on (Fig.~\ref{fig:design parts}(10)). In a simplified model, the active surface can be considered as a belt rotating around an angled axis of the CASM. The drawback of this angled design is the increased footprint that comes with larger angles. However, the design is parameterized so that the angle is easily changeable based on the hand used. If only two RRs are used, the active surfaces of the two RRs arranged orthogonally will maximize the manipulability of the system. For the proof of concept, we empirically selected a $30^{\circ}$ angle between the active surface and CASM to balance the performance and form factor of the design. Further discussion of this theorem can be found in Sec.~\ref{sec:motionModel} and evaluations in Sec.~\ref{sec:experiments}. 

\begin{table}[t]
    \centering
    \caption{Model Specifications of the Roller Ring (Labels can be found in Fig.~\ref{fig:design parts},~\ref{fig:CASM variants})}
    \label{tab:model specifications}
    \resizebox{0.45\textwidth}{!}{%
    \begin{tabular}{|c|c|c|}
        \hline
        \textbf{Symbol} & \textbf{Description} & \textbf{Value} \\
        \hline
        \textit{$d_{\text{ca}}$} & Distance to Contact Area [mm] & 6 \\
        \textit{$w_{\text{belt}}$} & Width of Timing Belt [mm] & 18 \\
        \textit{$t_{\text{belt}}$} & Thickness of Timing Belt [mm] & 0.5 \\
        \textit{$l_{\text{bp}}$} & Length of Backpack [mm] & 20 \\
        \textit{$w_{\text{o}}$} & Outer Width of CASM [mm] & 17 \\
        \textit{$w_{\text{ih}}$} & Inner Width of Human CASM [mm] & 12 \\
        \textit{$w_{\text{io}}$} & Inner Width of Model O CASM [mm] & 8 \\
        \hline
    \end{tabular}%
    }
\end{table}

\section{Motion Model}
\label{sec:motionModel}

In this section, we derive a general motion model for active surface-based in-hand manipulation. We show that a complete in-hand manipulation skill set can be provided by as few as $2$ active surface contacts through non-holonomic object motions, and that more active surface contacts can further improve the manipulation flexibility. We start with deriving a motion model that focuses on the in-hand rotation of a unit sphere, and then extend this model to the in-hand rotation of arbitrary object shapes. Thereafter, we derive an in-hand translation motion model, which can be combined with the rotation motion model to provide a complete set of in-hand manipulation solutions.

\subsection{In-hand Rotation of A Unit Sphere}
\begin{figure}[t]
    \centering
    \includegraphics[width=1.0\linewidth]{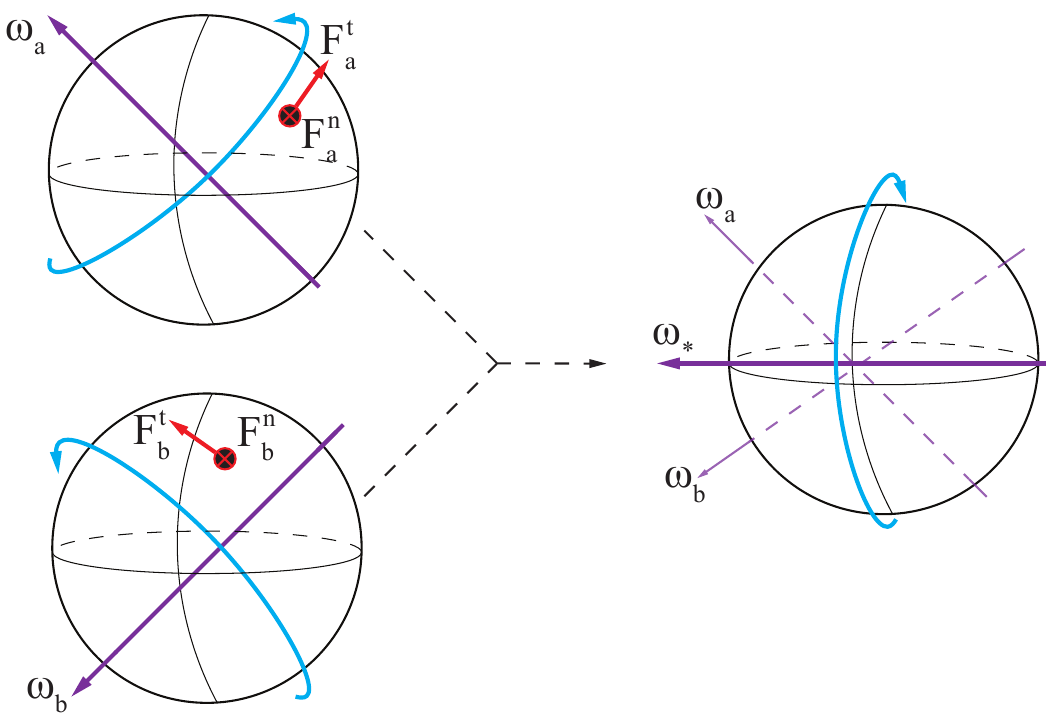}
    \caption{In-hand rotation motion model for a unit sphere visualized with $2$ active surface contacts. \emph{Left:} Angular velocities $\omega_a$ and $\omega_b$ independently actuated by active surfaces at points $p_a$ and $p_b$. \emph{Right:} The object's angular velocity is determined by the weighted sum of $\omega_a$ and $\omega_b$ as given by Eq.~\eqref{eq:modelRotation1}. Shown is an example where both contacts have the same weight.}
    \label{fig:single_sphere_model}
\end{figure}

Let us begin by assuming the manipulated object is a unit sphere object $\mathbb{S}^2$. By making a contact with an active surface at a point $p_a \in \mathbb{S}^2$, as illustrated in Fig.~\ref{fig:single_sphere_model}, the linear motion of the active surface at this contact will generate a torque to this sphere with respect to its center: $\tau_{a}=p_a \times F^t_a$
where $F^t_a \in \mathbb{R}^3$ is the tangential friction force at point $a$, as determined by the contact normal force $F_a^n \in \mathbb{R}^3$ and the friction coefficient $\mu \in \mathbb{R}$: $|F^t_a| \le \mu |F^n_a|$. Note that, as it will be discussed later, depending on the \emph{relative linear motion} between the active surface and the sphere at $p_a$, the direction of the force $F^t_a$ can vary, but the magnitude will stay constant when the two contacting surfaces are \enquote{slipping or scratching} against each other at these contacts.

With this single active surface contact that moves with a linear velocity $v_a \in \mathbb{R}^3$, the sphere object will accelerate to synchronize its motion at the point $p_a$ until the two surfaces in contact do not slip, resulting in an angular velocity of the object $\omega_a$ that satisfies: $v_a = \omega_a \times p_a$
where $\omega_a = \hat{\omega}_a \dot{\theta}_a \in \mathbb{R}^3$ is expressed by an angular velocity $\dot{\theta}_a \in \mathbb{R}$ and an axis of rotation $\hat{\omega}_a = \frac{\omega_a}{|\omega_a|}$. Similarly, if there is a single active surface contact $p_b \in \mathbb{S}^2$ that moves at a speed of $v_b \in \mathbb{R}^3$, the angular velocity caused by this contact alone is $\omega_b$ that satisfies $v_b = \omega_b \times p_b$. Both cases are exemplified in Fig.~\ref{fig:single_sphere_model}.

Now, if both contacts $p_a$ and $p_b$ are made with active surfaces, the sphere object will then initially start rotating with \enquote{slipping} contacts. However, unless the two active surface motions are colinear (in terms of the rotation axes), the resultant or converged motion at equilibrium will still be \enquote{slipping} to balance between the generated torque differences. To derive the object motion model based on the motions of these two active surface contacts, our goal is to find the angular velocity $\omega_*$ of the object when all \enquote{scratching or slipping} forces are balanced. The linear velocity of the sphere object at the contacts are:
\begin{equation}
    v_a^* = \omega_* \times p_a, \;\;\;\;\;\; v_b^* = \omega_* \times p_b,
\end{equation}
which are different from the linear velocities $v_a$ and $v_b$ of the two active surface contacts. As discussed above, this will cause \enquote{slipping motions} at both contacts, where the motions of the active surfaces relative to the object are given by $v_a - v_a^*$ and $v_b - v_b^*$. Such slippage will result in tangential forces at the contacts aligned to the relative motion directions:
\begin{equation}
    \begin{aligned}
        &\tau_a^* = p_a \times (\mu |F_a^n| \frac{v_a - v_a^*}{|v_a - v_a^*|}) = k_a p_a \times (v_a - v_a^*)\\
        &\tau_b^* = p_b \times (\mu |F_b^n| \frac{v_b - v_b^*}{|v_b - v_b^*|}) = k_b p_b \times (v_b - v_b^*)
    \end{aligned}
    \label{eq:derivedTorque}
\end{equation}
where $k_a = \frac{\mu |F_a^n|}{|v_a - v_a^*|}$ and $k_b = \frac{\mu |F_b^n|}{|v_b - v_b^*|}$ are scalar scaling factors determined by the magnitudes of both the contact forces and the relative velocities. When the object motion is at equilibrium (i.e. there is zero effective torque applied by the contacts) its motion will stabilize and the torques at both contacts should satisfy $\tau_a + \tau_b = 0$, yielding:
\begin{equation}
        k_a p_a \times (v_a - v_a^*) = -k_b p_b \times (v_b - v_b^*)
\end{equation}

since $\omega = p \times v$, we have:
\begin{equation}
    k_a (\omega_a - \omega_*) = -k_b (\omega_b - \omega_*)
\end{equation}
where $\omega_a$ and $\omega_b$ are the angular velocities independently generated by the contacts. Rearranging the former, we obtain the expression of the object's angular velocity:
\begin{equation}
    \omega_* = \frac{k_a \omega_a + k_b \omega_b}{k_a + k_b}
    \label{eq:modelRotation1}
\end{equation}

From Eq.~\eqref{eq:modelRotation1}, we can see that the angular velocity of the sphere object, as simultaneously actuated by 2 active surfaces, is simply a weighted sum of the object's angular velocities when actuated by each contact independently. Interestingly, the weights $k_a$ and $k_b$ are determined by the contact force and the intrinsic velocities of the active surfaces. In other words, the difference between the velocities of the active surfaces will directly determine the motion of the object $\omega_*$, making the motion model a \emph{differential motion model}. Intuitively, we know that if we want to rotate the object with motions more biased by a contact, it can be done by changing the contact force or speed. Fig.~\ref{fig:single_sphere_model} shows an example of when two contacts are simultaneously actuated by active surfaces at the same velocity and force.

Additionally, assuming non-colinear configurations of the active surfaces, we can see that the object's motion can be activated to rotate about multiple, essentially infinite, number of axes of rotation. As proven in \cite{morgan2022complex}, as long as a hand-object system can rotate an object about at least $2$ orthogonal axes, the object can be reoriented from any angle to any other angle. Although the set of axes of rotation generated by Eq. ~\eqref{eq:modelRotation1} cannot guarantee to contain orthogonal axes, the generated axes can be projected to provide rotations about orthogonal pairs of axes

Therefore, it is evident from Eq.~\eqref{eq:modelRotation1} that with as few as $2$ active surfaces, it can provide a \emph{complete} rotational manipulation solution. However, because of the lack of guarantees for a pair of orthogonal axes with so little active surfaces, the generated motions are not necessarily able to travel in arbitrary directions of rotation. Thus, the corresponding rotation-based manipulation solution is a \textit{non-holonomic} motion system that necessitates the object to often take “detours” from its current orientation to reach its goal orientation given the lack of aforementioned orthogonal guarantees.

Furthermore, it is straightforward to extend Eq.~\eqref{eq:modelRotation1} to $N$ active surface contacts. We omit the derivation steps here, and directly write down the motion model:
\begin{equation}
    \omega_* = \frac{\sum_{i=1}^N k_i \omega_i}{\sum_{i=1}^N k_i}
    \label{eq:modelRotation2}
\end{equation}
With more active surface contacts, Eq.~\eqref{eq:modelRotation2} will be able to provide more axes of rotation, reducing the \enquote{detours} taken by the object to reach its goal. The application of this principle can be seen in Sect. \ref{sec:experiments}.

\begin{figure}
    \centering
    \includegraphics[width=1.0\columnwidth]{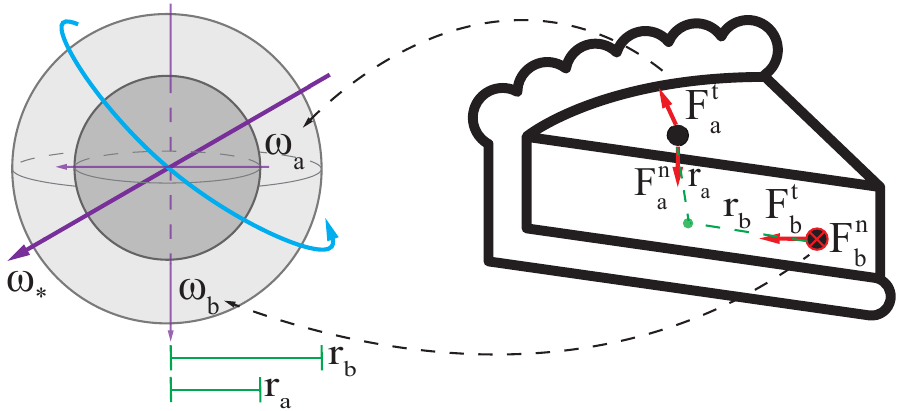}
    \caption{Illustration of the multi-sphere model for rotating arbitrary objects in-hand with active surface contacts. \emph{Right:} two active surface contacts $p_a$ and $p_b$ on a pie-shaped object. \emph{Left:} an equivalent view of the physical effect of these two contacts visualized on two virtual spheres of radii $r_a, r_b$.}
    \label{fig:multi_sphere_model}
\end{figure}

\subsection{Multi-Sphere Motion Model for Arbitrary Object Shapes} 
Extending from the unit-sphere motion model, we can now model the rotations of arbitrary object shapes as actuated by active surface contacts. We begin with $2$ contacts $p_a$ and $p_b$. However, this time the contacts are not necessarily on a unit sphere $\mathbb{S}^2$. Their distances to the axis of rotation are represented by $r_a, r_b \in \mathbb{R}$. Intuitively, these two contacts are actuating the object as if there were $2$ different contactable spheres, as illustrated in Fig.~\ref{fig:multi_sphere_model}. Thus, we know that the torque generated by these differing unit spheres are not weighted the same given the difference in radii. As such, the torques generated by these two contacts, as shown in Eq.~\eqref{eq:modelTorques}, are now scaled by $r_a, r_b$ to account for this difference:
\begin{equation}
    \begin{aligned}
        &\tau_a^* = r_a k_a p_a \times (v_a - v_a^*)\\
        &\tau_b^* = r_b k_b p_b \times (v_b - v_b^*)
    \end{aligned}
    \label{eq:modelTorques}
\end{equation}

As such, similarly to the derivation of the unit sphere motion model, the torques will balance out when the object is rotating at an equilibrium state. Skipping the intermediate steps, we can obtain the rotation motion model for $N$ active surface contacts on arbitrary objects of non-uniform size:
\begin{equation}
    \omega_* = \frac{\sum_{i=1}^N r_i k_i \omega_i}{\sum_{i=1}^N r_i k_i}
    \label{eq:modelRotation3}
\end{equation}
Based on this multi-sphere motion model, Eq.~\eqref{eq:modelRotation3}  provides a \emph{complete and non-holonomic} rotational manipulation solution. However,  a major difference now is that the individual angular velocities are additionally weighted by the distances between the contact and the rotation axis, providing another dimension for rotations to be regulated by contact variances.

\subsection{Motion Model for In-hand Translations}
To achieve a complete in-hand manipulation solution, the final component required in the motion model is the ability to translate the object. The model analysis will focus on pure translations supplementary to the derived rotation models. In practice, however, rotation and translation can happen simultaneously during manipulation in order to perform more efficient actions. 

For pure in-hand object translation, the first requirement is that the object's angular velocity should be zero. Based on our model above, we know that this can be expressed as a motion constraint among all active surfaces: $\omega_* = \sum_{i=1}^N r_i k_i \omega_i = 0$. When this constraint is satisfied, which can be achieved in an infinite number of ways when multiple contacts exist to cancel out each other’s torques, the velocity of the object is purely determined by the linear forces provided by the active surfaces. Let us denote by $v^* \in \mathbb{R}^3$ the linear velocity of the object, by $v_i \in \mathbb{R}^3$ and $F_i^t$ the linear velocity and the tangential force at the $i$-th active surface. The following equation further constrains the object's translation:
\begin{equation}
    \sum_{i=1}^N F_i^t \cdot \frac{v_i - v^*}{|v_i - v^*|} \cdot \frac{v_i \times v^*}{|v_i \times v^*|} = 0
    \label{eq:modelTranslation}
\end{equation}

In Eq.~\eqref{eq:modelTranslation}, the tangential force at each contact is first aligned to the direction of the \enquote{slipping or scratching} motion and then projected to the orthogonal direction of the object's motion by dot product. The zero-sum of all these forces guarantees that the object is purely translating in the direction of $v^*$. In practice, although it is not difficult to obtain $\omega_*=0$, as discussed above, achieving arbitrary translation motions is challenging due to the additional constraint (Eq.~\eqref{eq:modelTranslation}), especially when the number of contacts, $N$, is small. Nonetheless, given our comprehensive solution of in-hand rotation, we will be able to manipulate the object from any pose to any other by sequencing translation and rotations. Thus, a manipulation graph can be constructed to generate such action sequences\cite{cruciani2018dexterous}, although the aforementioned \enquote{detours} are not avoidable due to the relatively small number of contacts wherein increasing the number of non-colinear contacts can reduce but not entirely remove the degree of detours necessary.
\section{Experiments}
\label{sec:experiments}

\begin{figure}
    \centering
    \includegraphics[width=1\linewidth]{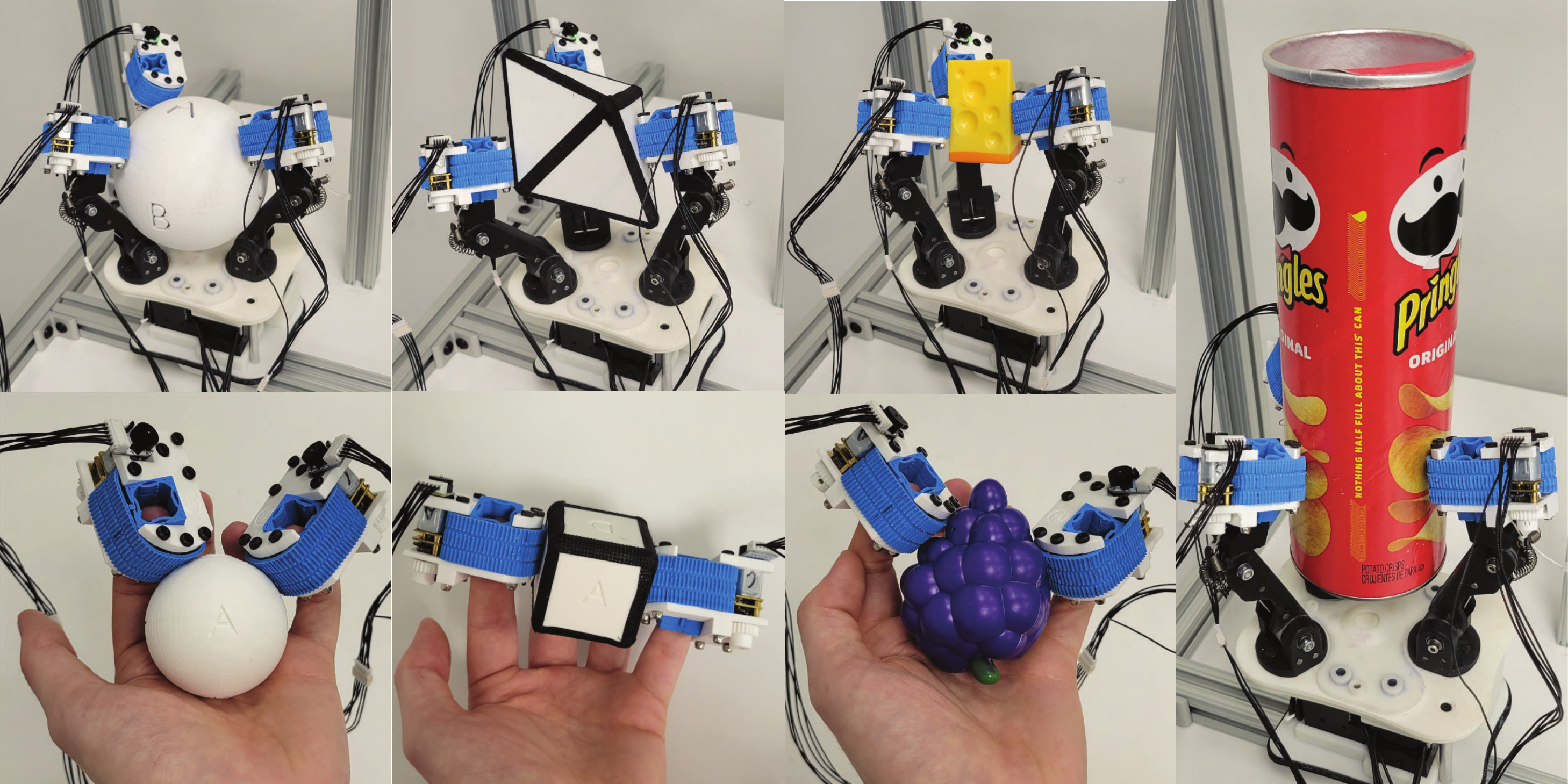}
    \caption{Manipulated experimental objects(Top Row Left to Right: A-F Sphere, Octahedra, Children's Cheese Toy, Cardboard Tube; Bottom Row Left to Right: A-F Sphere, Cube, Children's Grape Toy) } 
    \label{fig:experiment objects}
\end{figure}

To evaluate the design and motion model of the RR, we constructed a prototype, as seen in Fig.~\ref{fig:design parts} to test its in-hand manipulation capabilities using various objects (Fig.~\ref{fig:experiment objects}). The RRs were affixed to both a Yale Model O hand and a human hand for all manipulation tasks. Each RR had its velocity set to a differing set of values during each experiment to generate differential motions (Sec.~\ref{sec:motionModel}) and controlled via velocity control teleoperation. No retuning was required between each hand and the same controller was used across all experiments. Execution times were measured by analyzing motion video frames from the start to stop of motions. For each object, the reported time represents the average of six successful trials.

\subsection{Rotational Manipulation}
\begin{figure*}[t]
    \centering
    \includegraphics[width=1.0\linewidth]{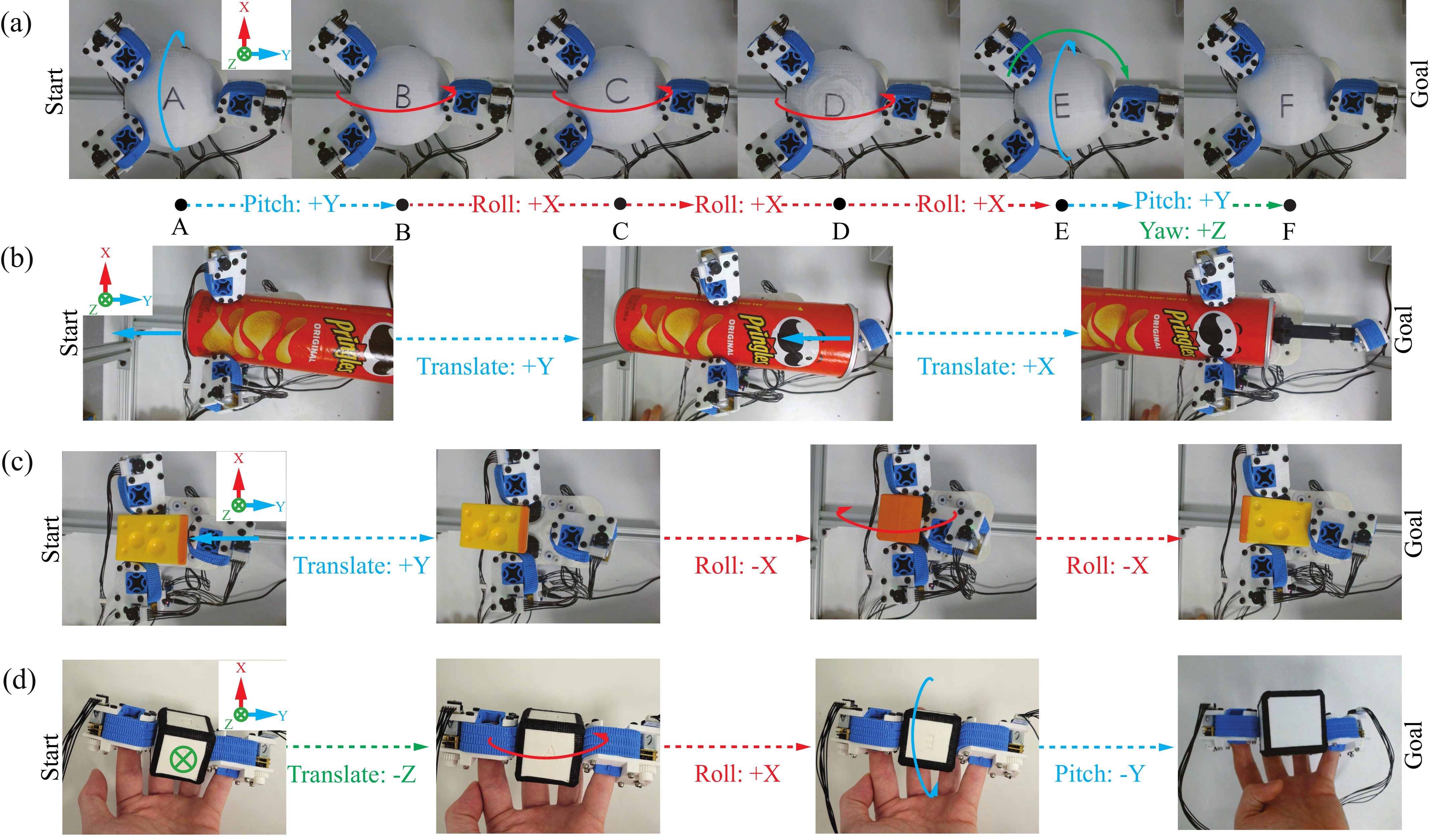}
    \caption{Motion Examples of the RR: (a)A-F Sphere reoriented by the RR through all faces when held in a power grasp by the Yale Model O (b) Cardboard tube translated along the Yale Model O when held in a power grasp (c) Cheese toy translated and rotated when held in a precision grasp by the Yale Model O (d) Cube translated and rotated when held in a precision grasp by a human hand} 
    \label{fig:motion example}
\end{figure*}

In these experiments, we evaluated the objects' ranges of motion in orientations that can be imparted by the RRs. For all situations in which the manipulated object's geometry did not collide internally and a stable grasp was maintained, we aimed to confirm that the designed RRs could rotate a grasped object from any initial orientation to any goal orientation to verify the theory derived in Sec.~\ref{sec:motionModel}. Based on the single-sphere motion model with $3$ RRs, the sphere could be rotated to any face relatively quickly (averaging 23.48$s$ traveling through all faces) about multiple axes (Fig.~\ref{fig:motion example}(a)) as opposed to the limited differential motions provided by $2$ RRs (Sec.~\ref{sec:motionModel}). With the single-sphere model verified, we moved towards the verification of the multi-sphere model. To do so, we chose a cheese toy and that found that the toy could also be rotated to a desired orientation robustly in approximately 10.98$s$ (Fig.~\ref{fig:motion example}(c)). From this, we can verify the in-hand rotation capability of the RR design and multi-sphere motion model. Between these two objects, additionally, we can note that the difference in timing is nominally due to both the object size as well as necessary detours for the manipulation.

As discussed by the motion models and visualized in the supplementary video, \enquote{detours} were often needed for in-hand rotations since the RRs can only provide differential non-holonomic manipulation. Supplementing more RRs would reduce and/or negate these detours, improving the overall manipulation timing.

\subsection{Translational Manipulation}
Utilizing the translational portion of the Motion Model (Sec.~\ref{sec:motionModel}) we were able to generate translations along all cardinal planes from the base set of RRs. Following the same constraints for rotational manipulation, as long as the object did not collide internally and is stably grasped, we found that translations were easily generated. As seen in Fig.~\ref{fig:motion example}(b-d), translations were generated for grasped objects of varying geometry such as a Pringles tube, cheese toy, and cube. We observed that, similarly to the rotational manipulation, the type of hand made no difference in terms of manipulation capabilities. Translations can be generated in sequences to fully manipulate the in-hand object from any pose to any others, as discussed in Sec.~\ref{sec:motionModel}.

However, as previously discussed, these manipulation paths often take \enquote{detours} to fulfill these tasks. Despite this, translations had no adverse affect on manipulation task timing, as the Pringles can and cube translation could be performed in approximately 6.78$s$ and 9.87$s$ respectively; thus we found that rapid translation manipulation tasks were feasible with as little as $2$ RRs.

\subsection{Limitations of Roller Rings}
1) \textit{Geometric Constraint:} Due to the geometry of the RR, objects  larger than the hand's power grasp task-space or thinner than the formed gap are difficult to manipulate. These objects can be manipulated when held in a precision grasp, but with a decrease in stability. Utilizing the palm and fingers as a support improves the in-hand manipulation, but is not effective due to the lack of active surfaces. Future analysis on RR manipulation on certain hand sizes can be performed to better understand its capabilities within this hand-object system.

2) \textit{Friction:} Because the angle of the CASM was fixed to $30^{\circ}$ to limit the form factor, the RRs require \enquote{slipping and scratching} motions orthogonal to its main motion axes for manipulation. Such motions will not only make the system less energy-effective, but can also reduce, the system's durability due to constant \enquote{scratching}. Further experimentation is required to understand and denote the trade-off between lost energy and manipulation timing.

3) \textit{Non-Compliance:} The RRs are not compliant, thus, contacts between the grasping system and object are difficult to maintain using the RRs alone. Intuitively we can ascertain that because compliant mechanisms can conform to an object's geometry, the contact generated by the RR will be more stable. Thus, in order to facilitate the manipulation stably, the level of compliance achievable by the RR is directly correlated to the compliance of the underlying gripper. Therefore, RRs on a non-compliant system will encounter difficulties performing in-hand manipulation, as the contacts will be easily broken. By either increasing or having some base level of compliance in the base gripper, the non-compliancy of the RR can be overcome as mirrored in our experimentation of the RR. Compliant grasping systems, such as the Yale Model O or human hand, were required in order to perform stable in-hand manipulation due to their ability to inherently push the RR towards the surface of the object. Because of this, we saw that the human hand had a smaller number of failed trials due to its higher level of compliancy. 

\section{Conclusion and Future Work}
\label{sec:conclusion}
This paper proposed the design of a wearable Roller Ring, utilizing the concept of active surfaces, that can be used to enhance the in-hand manipulation capabilities of existing grasping systems. Active surface-based motion models were derived to show that active surface-based manipulation provides a differential and non-holonomic manipulation model. The motion model also proves that the proposed design is able to provide a complete manipulation skill set, including in-hand rotations and translations. Experiments were performed to physically verify the manipulation capabilities of RRs. Different objects of varying sizes and geometries were manipulated to empirically validate the applicability of our design in daily manipulation tasks. In future work, we plan to mechanically make the RRs more compliant, as well as reduce the differential motion friction. Based on the derived motion models, we will develop algorithms to provide general and optimized manipulation solutions.

\bibliographystyle{IEEEtran}
\bibliography{reference}

\end{document}